\def\year{2018}\relax
\documentclass[letterpaper]{article} %DO NOT CHANGE THIS
\usepackage{aaai18}  %Required
\usepackage{times}  %Required
\usepackage{helvet}  %Required
\usepackage{courier}  %Required
\usepackage{url}  %Required
\usepackage{amsmath}
\usepackage{amssymb}
\usepackage{graphicx}  %Required

\newcommand*\samethanks[1][\value{footnote}]{\footnotemark[#1]}

\DeclareMathOperator*{\argmax}{arg\,max\,}

\newcommand{\realset}{\mathbb{R}}
\newcommand{\bc}{{\bf c}}
\newcommand{\bu}{{\bf u}}
\newcommand{\bw}{{\bf w}}

\newcommand{\bx}{{\bf x}}

\newcommand{\bM}{{\bf M}}
\newcommand{\cN}{\mathcal{N}}

\begin{document}
\title{Customized Nonlinear Bandits for Online Response Selection in \\ Neural Conversation Models}
\author{Bing Liu\thanks{These two authors contributed equally.}, Tong Yu\samethanks, Ian Lane, Ole J. Mengshoel\\
Electrical and Computer Engineering, Carnegie Mellon University\\
\{liubing, lane\}@cmu.edu, tongy1@andrew.cmu.edu, ole.mengshoel@sv.cmu.edu\\
%()
}
\maketitle
\begin{abstract}
    Dialog response selection is an important step towards natural response generation in conversational agents. Existing work on neural conversational models mainly focuses on offline supervised learning using a large set of context-response pairs. In this paper, we focus on online learning of response selection in retrieval-based dialog systems. We propose a contextual multi-armed bandit model with a nonlinear reward function that uses distributed representation of text for online response selection. A bidirectional LSTM is used to produce the distributed representations of dialog context and responses, which serve as the input to a contextual bandit. In learning the bandit, we propose a customized Thompson sampling method that is applied to a polynomial feature space in approximating the reward. Experimental results on the Ubuntu Dialogue Corpus demonstrate significant performance gains of the proposed method over conventional linear contextual bandits. Moreover, we report encouraging response selection performance of the proposed neural bandit model using the Recall@k metric for a small set of online training samples.
\end{abstract}

\section{Introduction}
    Conversational agents, or chatbots, have a wide range of applications such as in technical support, personalized service, and entertainment~\cite{young2013pomdp,serban2016building}. Conventional approaches in conversational modeling include using rule-based and learning-based methods \cite{oh2000stochastic,schatzmann2006survey} that typically require expert knowledge in designing rules and features. Such systems are usually designed to model dialogs in a specific domain. Transferring such models from one domain to another is often difficult.

    With the explosive growth of conversational data in social media, neural network based data-driven approaches \cite{shang2015neural,vinyals2015neural} to conversational modeling have been explored. These models are trained based on large dialog corpora, making few assumptions about dialog domain or structures. They are suitable in modeling non-task-oriented conversations, and have shown promising performance in modeling social media chats \cite{ritter2011data,shang2015neural,li2016persona} and movie conversations \cite{vinyals2015neural,li2016deep}. Most of the existing work focuses on the offline supervised dialog model training setting, in which the models are trained with a large number of context-response pairs. Online learning methods for neural conversational models, which are useful in building personalized dialog systems and transfering systems to new dialog domains, is not widely studied in literature. 

    In neural conversational models, response to a dialog context can either be generated directly by the model or can be selected from a repository of predefined responses. Generation-based models have advantages in producing responses that are more diverse. However, they often require a large number of training samples. Evaluating generation-based systems is also challenging, with recent studies \cite{liu2016not} showing that many of the automatic evaluation methods for response generation are poorly correlated with human judgement. Retrieval-based models that select a dialog response from a candidate set based on a dialog context can be seen as an intermediate step towards generation-based systems. Response selection is similar to response generation when the set of candidate responses is very large \cite{lowe2015ubuntu,lowe2016evaluation}. There are well defined automatic evaluation metrics for response selection models.

    In this work, we focus on online learning for retrieval-based conversational models. We propose a neural nonlinear bandit model (NNBM) that can be trained online for dialog response selection. Given a dialog context, the model selects a most likely response from a list of response candidates. Binary feedback is collected from users indicating whether or not the selected responses are appropriate. This binary feedback serves as the reward to the contextual bandits. 

    Our main contributions in this work are:
    \begin{itemize}
        \item We design a contextual bandit model, NNBM, that has a nonlinear reward function applied on selected dimensions of a polynomial feature space. The proposed method significantly outperforms contextual bandits with linear rewards that are widely used in online recommendations.
        \item We apply neural contextual multi-armed bandits to online learning of response selection in retrieval-based dialog models. To our best knowledge, this is the first attempt at combining neural network methods and contextual multi-armed bandits in this setting.
    \end{itemize}

\section{Related Work}
\label{related_work}

    \begin{figure}[t]
        \centering
        \includegraphics[width=220pt]{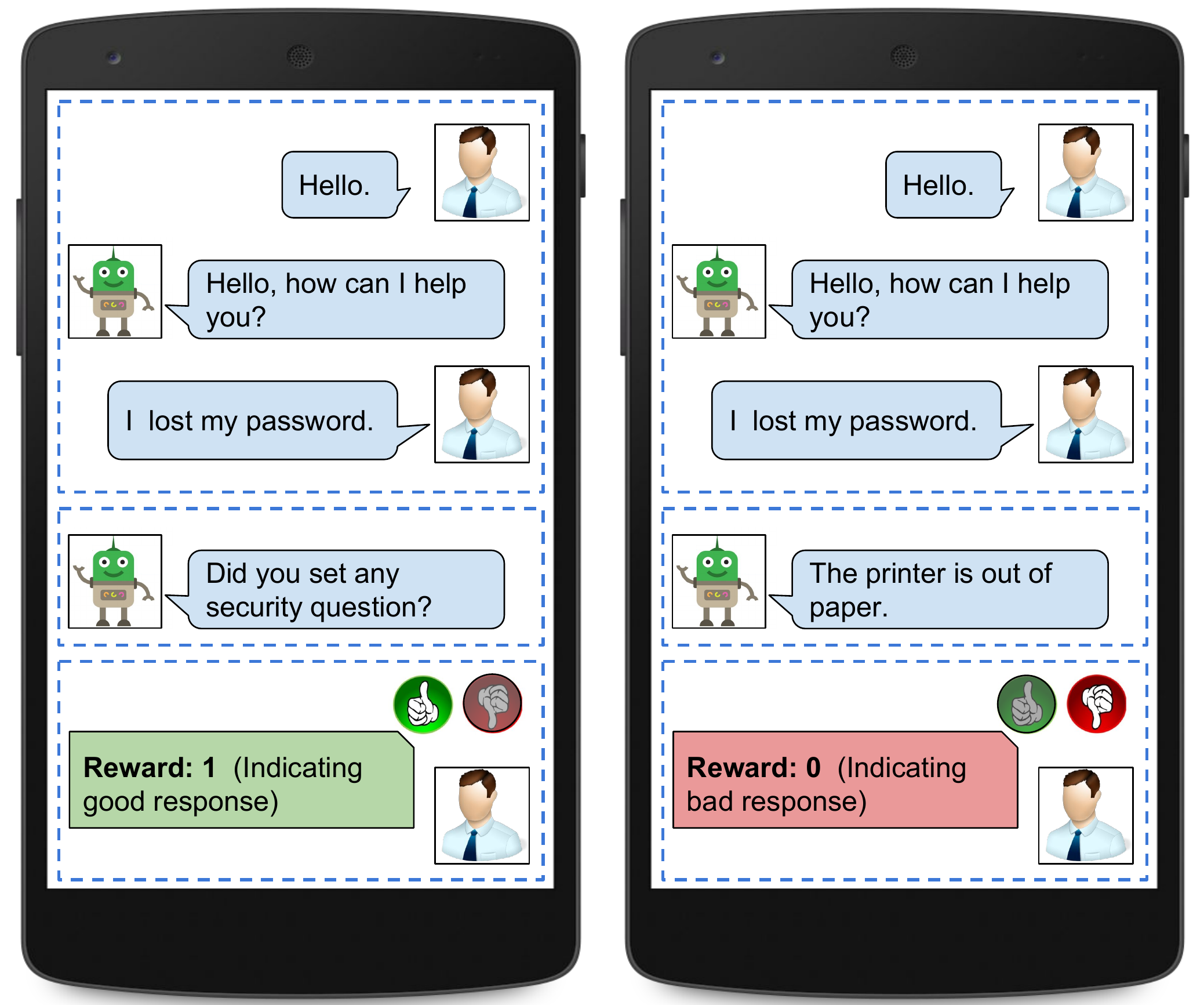}
        \caption{Illustration of the online model for response selection in dialog systems. Dialog history with a few number of turns is provided to the agent as the dialog context. The agent computes the most likely true response from a list of candidate responses and returns it to user. The user then provides binary feedback, positive (left) or negative (right), to the agent indicating the quality of the returned response. }
        \label{fig:Dialogue_Bandit}
    \end{figure}

\subsection{Conversation Modeling} 
    Traditional approaches to conversation modeling typically involve template and rule based methods with statistical learning components ~\cite{levin1997stochastic,oh2000stochastic}. With the growth of conversational data in social media, data-driven approaches to conversation modeling have been widely studied. Ritter et al.~\shortcite{ritter2011data} framed conversation response generation as a statistical machine translation problem. Vinyals et al. ~\shortcite{vinyals2015neural} designed a neural conversation model that can generate simple conversations and extract knowledge from an open-domain dataset. Shang et al.~\shortcite{shang2015neural} showed that the encoder-decoder-based neural network model could generate varied multiple responses to a given post on social networks. Serban et al. ~\shortcite{serban2016building} proposed a hierarchical recurrent neural model that utilizes context over an extended dialog history. Li et al. ~\shortcite{li2016diversity} proposed using maximum mutual information as the objective function to produce more diverse and interesting responses. These end-to-end neural conversation models are trained offline with supervised learning using large parallel corpora of utterance-response pairs. In this work, we focus on online learning of neural conversational models. 

    Reinforcement learning (RL) has also been applied in dialog modeling. In task-oriented dialog systems, RL can be applied to optimize dialog policy online with the feedback collected via interacting with users \cite{singh2000empirical,gavsic2013line}. Zhao and Eskenazi \cite{zhao2016towards} introduced joint learning of dialog state tracking and dialog policy learning using deep RL. Li et al. \cite{li2017end} proposed end-to-end trainable task-oriented dialog model with deep RL. Lipton et al. ~\shortcite{lipton2016efficient} proposed a Bayes-by-Backprop Q-network that effectively improves the efficiency of exploration with Thompson sampling. He et al. ~\shortcite{he2015deep} proposed a deep reinforcement relevance network (DRRN) that applies to text-based games in which action and state spaces are represented with separate embedding vectors and used to approximate the Q-function. These models are applied in a sequential decision problem setting for task-oriented dialogs. Our model, on the other hand, focuses on online learning in a setting where the context is provided and is not dependent on previous state and actions. The problem setting is closer to a non-task-oriented chit-chat dialog setting. In non-task-oriented dialog learning, Li et al.~\shortcite{li2016deep} introduced the use of RL for response generation by simulating dialogs between two agents that optimizes future reward. Serban et al. \shortcite{serban2017deep} proposed an hybrid generation and retrieval based dialog model that applies RL in response selection. In these models, a simple reward function with linear approximation is applied in learning dialog policy. In this work, we investigate online learning efficiency using a more general nonlinear reward function approximation with Thompson sampling. 
    
    Evaluation of dialog system is difficult \cite{liu2016not} as there is not yet a suitable metric for response generation evaluation. Many recent studies~\cite{lowe2015ubuntu,kadlec2015improved} using the Ubuntu Dialogue Corpus instead proceed towards best response selection, which can be seen as an intermediate step towards response generation~\cite{lowe2015ubuntu} and has well accepted evaluation metrics. We use the same response selection methods and metrics in our study.

\subsection{Bandits and Online Learning}
    Multi-armed bandit model is a popular online learning method with great success in many fields, such as online advertisement \cite{chapelle2011empirical}, recommender systems \cite{li2010contextual}, and stochastic combinatorial optimization \cite{gai12combinatorial,kveton14matroid}. The bandit enables online model update online based on the feedback from the environment, which has several nice properties in terms of finite-time regret bounds with a clear dependence on the parameters of interest. To model more complicated problems, a number of non-linear bandit models \cite{filippi2010parametric,gopalan2014thompson,kawale2015efficient,katariya2017stochastic,ijcai2017-278,yu2017thompson} are proposed, in which the rewards are approximated by different non-linear functions. Bandits have also been applied in some dialog systems. 
    Genevay et al. \shortcite{genevay2016transfer} applied bandits in source user selection for user adaptation in spoken dialog systems. Bouneffouf et al. \shortcite{bouneffouf2014contextual} proposed active Thompson sampling method with bandits to select the most useful unlabelled examples to train a predictive model. They used a linear regression model to approximate the reward in each round of interaction. In our problem setting, the reward is a Bernoulli variable, where a logistic regression model is more suitable for binary reward approximation. Moreover, we apply bandits on distributed representation of the dialog context that are generated from neural network models, which is different from Bouneffouf et al.~\shortcite{bouneffouf2014contextual}.

    \begin{figure*}[t]
        \centering
        \includegraphics[width=420pt]{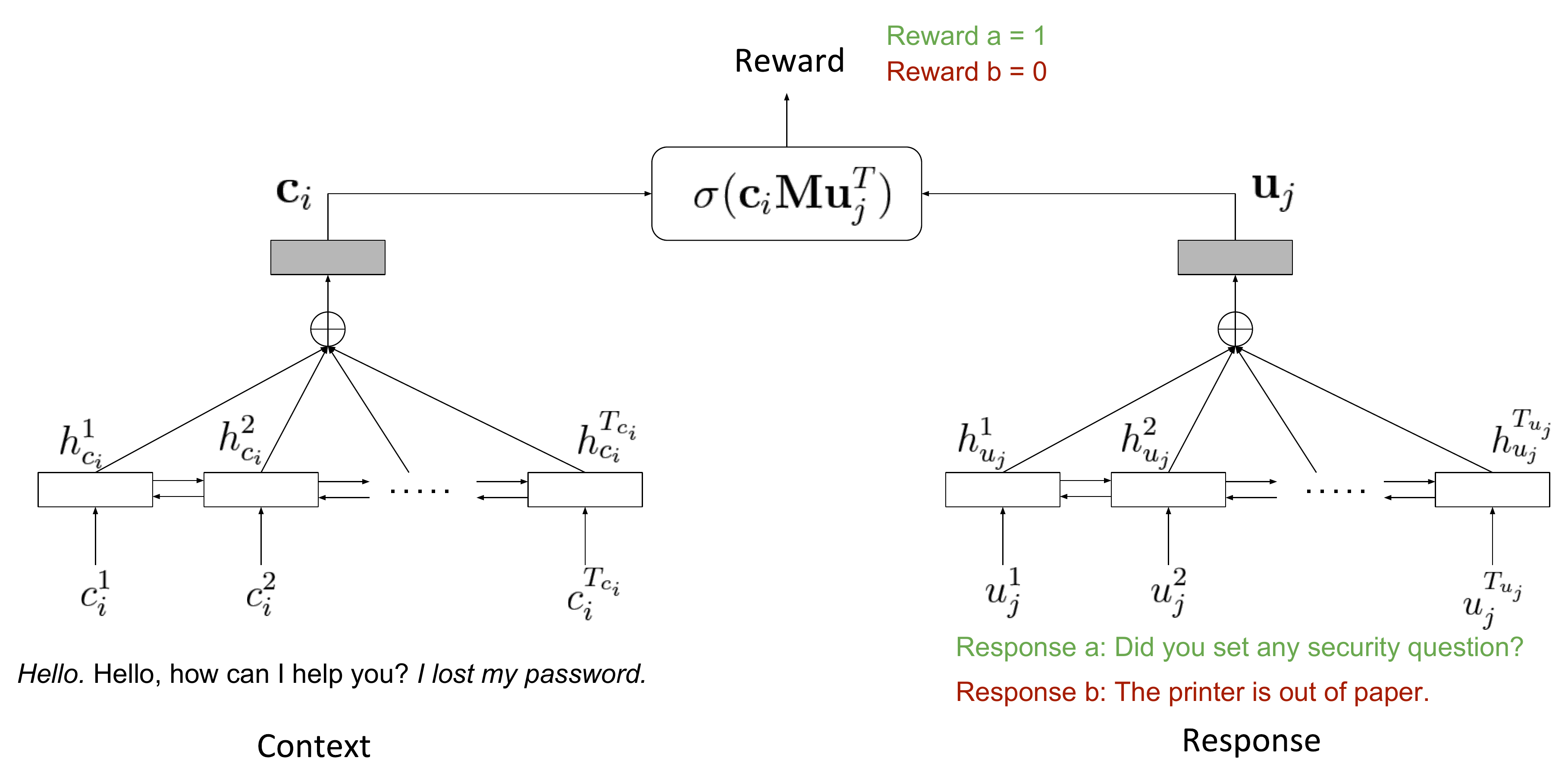}
        \caption{ Architecture of the proposed Neural Nonlinear Bandit Model (NNBM). A bidirectional LSTM is used to encode dialog context and response to continuous representations. The context and response representations $\textbf{c}_i$ and $\textbf{u}_j$ are produced by taking the average of the LSTM state output sequence. $\textbf{c}_i$ and $\textbf{u}_j$ are then mapped to a second order polynomial feature space and serve as the input to the contextual bandits. A binary reward is finally collected from user indicating the quality of the selected response to the dialog context, which is used to update the NNBM parameters. }
        \label{fig:rnn_bandit}
    \end{figure*}
    
\section{Neural Nonlinear Bandit Model}
    Figure \ref{fig:rnn_bandit} shows the architecture of the our proposed neural nonlinear bandit model for online response selection in retrieval-base dialog systems. Let $C = \{c_1, c_2, c_3 \dots\}$ represent a set of dialog context samples and let $U = \{u_1, u_2, u_3 \dots\}$ represent a set of response samples in the dialog corpus.\footnote{In this paper, we use $c_i$ to refer to the $i$th context sample and $\bc_i$ to denote the feature representation of the $i$th context sample. The same also applies to dialog responses.} A dialog context consists of a sequence of utterances by one or more users, and potentially by the NNBM. Given a dialog context $c_i$, the NNBM attempts to select a most likely response $u_j$ from a pool of candidate responses. We ask the user to provide a binary feedback indicating whether or not the selected response is an appropriate one. This binary value $r_{ij}$ serves as the reward to the contextual bandits during online learning. 
    
    We use a bidirectional LSTM to produce distributed representations of the dialog context and response. The context and response representations $\textbf{c}_i$ and $\textbf{u}_j$ are produced by taking the average of the LSTM sequence outputs $(h_{c_i}^{1}, ..., h_{c_i}^{T_{c_i}})$ and $(h_{u_j}^{1}, ..., h_{u_j}^{T_{u_j}})$. $\textbf{c}_i$ and $\textbf{u}_j$ serve as input features to the contextual bandit. In learning the bandit, we propose a Thompson sampling method that is applied to a second order polynomial feature space of the context and response in approximating the reward. We describe the proposed model details in the sections below.

\subsection{Utterance Encoding with LSTM}
\label{sec:nn}
    We use a bidirectional LSTM to generate distributed feature representations for both the context and the response. For the context reader, input to the network at each time step $t$ is the embedding of the word $c_{i}^{t}$ in context $c_i$. The corresponding output $h_{c_i}^{t}$ is the concatenation of the forward LSTM output $\overrightarrow{h}_{c_i}^{t}$ and backward LSTM output $\overleftarrow{h}_{c_i}^{t}$, i.e. $h_{c_i}^{t} = [\overrightarrow{h}_{c_i}^{t}, \overleftarrow{h}_{c_i}^{t}]$. The final context representation $\textbf{c}_i$ is produced by averaging over the sequence of LSTM state outputs. For the response reader, the same LSTM that shares the parameters with the context reader is used. The final response representation  $\textbf{u}_i$ for response $u_i$ is generated in the same manner as in the context representation generation. 

    The generated feature vectors $\textbf{c}_i$ and $\textbf{u}_i$ encode the meaning of the context and the response. During LSTM offline supervised pre-training, a score is generated for the context-response pair representing the probability of the match. The score of the context-response pair is calculated by applying the sigmoid function on the transformed inner product of $\textbf{c}_i$ and $\textbf{u}_j$ as:
    \begin{equation}
        p(\textup{label}_{i,j} = 1 | \textbf{c}_i, \textbf{u}_j) = \sigma (\textbf{c}_i \bM \textbf{u}_j^T)
    \end{equation} 
    where $\bM$ is a transformation matrix. In equation (1), the matrix $\bM$ is randomly initialized and updated during model training together with other neural network parameters. The neural network model is optimized by minimizing the cross entropy of all labeled context-response pairs in the training set:
    \begin{equation}
        \mathcal{L} = -\sum _{i,j} \log p(\textup{label}_{i,j} = 1| \textbf{c}_i, \textbf{u}_j)
    \end{equation} 
    
\subsection{Linear Logistic Regression Thompson Sampling}
\label{sec:m1}
    At round $t$ of interaction in online model learning, a random context $c_i$ is picked. Our agent chooses a user response $u_j$ and receives a binary reward (1 for like, and 0 for dislike). With context $c_i$ and response $u_j$, our algorithm applies a parametric approximation of the reward $r_{ij}$ by the representations $\bc_i$, $\bu_j$ and unknown model parameters $\bw$. As the reward in our problem setting is a Bernoulli variable, logistic regression is a suitable choice to model the reward.

    In online advertising or recommendations with contextual bandits \cite{li2010contextual,chapelle2011empirical}, the reward $r_{ij}$ is usually linear in feature representations of users, articles, or advertisements. In our case, we can also model $r_{ij}$ by a linear logistic function with parameters $\bw$, such that $r_{ij} = \sigma(\bx_t\bw)$, where $\bx_{t} = [\bc_i, \bu_j] \in \realset^{1 \times 2L}$, $\bc_i \in \realset^{1 \times L}$, $\bu_i \in \realset^{1 \times L}$ and $\sigma(.)$ is the sigmoid function.

    Many algorithms have been studied in literature for balancing exploration and exploitation during policy learning, such as exploit-only, random, Thompson sampling \cite{chapelle2011empirical}, and upper confidence bound (UCB) \cite{auer02finitetime}. We apply Thompson sampling as it outperforms UCB, exploit-only, and random methods \cite{chapelle2011empirical}.

    We first sample the parametrization of reward from its posterior. There is no closed-form solution for sampling from the posterior of logistic regression. Based on the Laplace's method, the posterior at time $t$ can be approximated \cite{spiegelhalter1990sequential,mackay1992evidence} by $\bw_t \sim \cN(\bar{\bw}_{t - 1}, S_{t - 1})\,$.The \emph{covariance matrix} is $S_t = (X_t^{T} C_t X_t + \lambda I_L)^{-1}$, where $C_t$ is a $t\times t$ diagonal matrix with diagonal elements $(C_t)_{ii} = \sigma( \bx_{i} \bar{\bw}_t)(1-\sigma(\bx_{i} \bar{\bw}_t))$. Here $X_t = (\bx_{1}^{T}, \dots, \bx_{t}^{T})^{T} \in \realset^{t \times 2L}$ is the matrix of all context and response pair $\bx_i$ up time $t$, where $1 \leq i \leq t$.

    To obtain a Gaussian approximation of the posterior distribution, we maximize the \emph{logistic regression log-likelihood} $L_t(\bw)$ from the first $t$ observations to get $\bar{\bw}_t$, which is the mean of the Gaussian. We update $t(\bw)$ incrementally according to Newton's method: 
    \begin{equation*}
        \bar{\bw}_t = \bar{\bw}_{t-1} - H_{t - 1}^{-1}(\bar{\bw}_{t-1})\nabla(L_{t - 1}(\bar{\bw}_{t-1})),
    \end{equation*}

    where $\bar{\bw}_{t - 1}$ is the solution at time $t - 1$, $H_t(\bw)$ is the Hessian of the \emph{logistic regression log-likelihood} $L_t(\bw)$ from the first $t$ observations, and $\nabla(L_t(\bw))$ is the corresponding gradient. The log-likelihood is defined as:
    \begin{equation}
    \label{eq:logregl}
    \begin{split}
         L_t(\bw) = &\lambda \bw^{T} \bw +
      \sum_{i = 1}^t  f_i(\bx_i) \log(\sigma(\bx_{i} \bw)) + {} \\ 
      &\sum_{i = 1}^t  (1 - f_i(\bx_i)) \log(1 - \sigma(\bx_{i} \bw)))\,,
      \nonumber
    \end{split}
    \end{equation}
    where $f_i(\bx_i)$ is the observed reward at time $i$ and $\lambda$ is the weight of the regularization term. After we sample $\bw_t$, we choose the response with the highest expected reward with respect to $\bw_t$:
    \begin{equation*}
      \bu_t = \argmax_{u_t \in \text{ all responses at $t$}} \sigma(\bx_t \bw_t)\,,
    \end{equation*}
    where $\bx_{t} = [\bc_t, \bu_t]$ and $\bc_t$ is the context at $t$.
    Finally, we observe $f_t(\bx_t)$ and receive it as a reward. 

\subsection{Logistic Regression Thompson Sampling on Polynomial Feature Space}
\label{sec:m2}
    We further propose a contextual bandit model with a nonlinear reward function applied on selected dimensions of a second order polynomial feature space. In approximating the reward, we apply the sigmoid function on the transformed inner product of $\bc$ and $\bu$, $\bc\bM\bu^{T}$:
    \begin{equation}
        \begin{bmatrix}
        \!\bc^{(1)}\! & \!\bc^{(2)} \! & \!\dots \! &\! \bc^{(L)\!} 
    \end{bmatrix}\!
        \begin{bmatrix}
       \! \bM_{11}\! & \!\bM_{12}\! & \!\dots \! &\! \bM_{1L} \\
       \! \bM_{21}\! &\! \bM_{22}\! & \! \dots \! & \!\bM_{2L} \\
       \! \vdots \!& \!\vdots\! &\!\ddots \!&\! \vdots \\
       \! \bM_{L1}\! & \!\bM_{L2} \!& \!\dots \! &\! \bM_{LL}
    \end{bmatrix}\!
        \begin{bmatrix}
        \!\bu^{(1)}\!\\
       \! \bu^{(2)}\!\\
       \! \vdots \!\\
       \! \bu^{(L)}\!
    \end{bmatrix}\!
    \end{equation}
    where $\bc^{(i)}$ and $\bu^{(j)}$ are the $i$th and $j$th dimension of $\bc$ and $\bu$. $\bM_{ij}$ indicates the element in the $i$th row and the $j$th column of $\bM$, which is also the weight coefficient of term $\bc^{(i)}\bu^{(j)}$. 

    This parametric approximation is similar to a degree-2 polynomial kernel function in the space of $\bx$, where $\bx = [\bc, \bu] \in  \realset ^{1 \times 2L}$. The only difference is that the weight coefficients of some terms are zeros. Let $\bx^{(i)}$ be the $i$th dimension of $\bx$. The coefficients of $(\bx^{(i)})^2$ and $\bx^{(i)}$ are zero where $1 \leq i \leq 2L$. Besides, the coefficients of $\bx^{(j)}\bx^{(k)}$ are zero where $1 \leq j \leq k \leq L$, or $L+1 \leq j \leq k \leq 2L$.
    The bias term's coefficient is zero as well. As a result, we approximate the reward by $\sigma(\phi(\bx)\bw)$, where $\phi(\bx)$ maps $\bx$ into a higher-dimensional space. In addition, those zero coefficients guide us to remove some useless terms but select the important ones in the explicitly mapped degree-2 polynomial space. Therefore, this method can provide more powerful approximation compared to linear approximation studied previously \cite{li2010contextual,chapelle2011empirical}.

    \begin{figure*}[ht]
      \centering
      \includegraphics[width=0.31\textwidth]{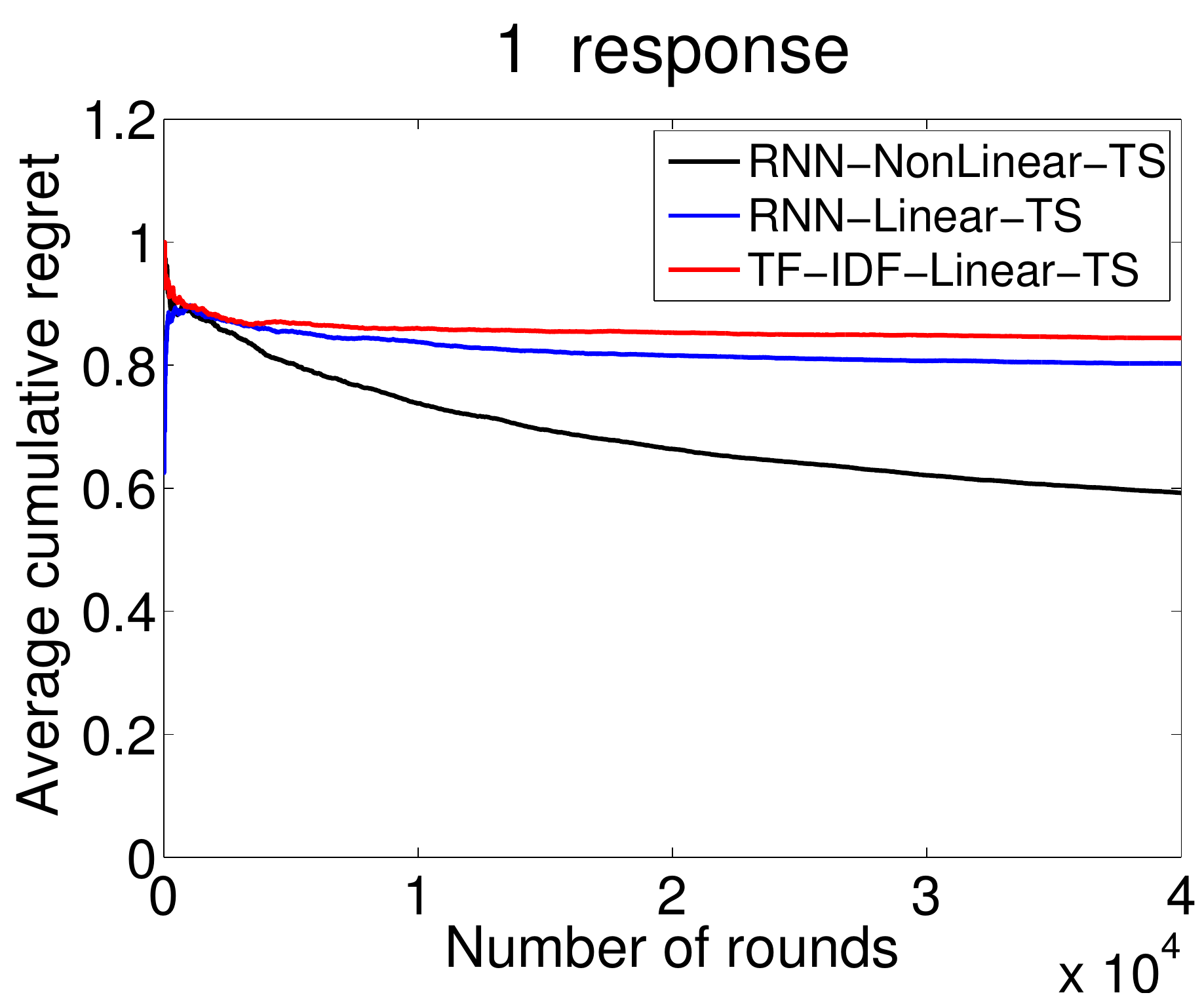}
      \includegraphics[width=0.31\textwidth]{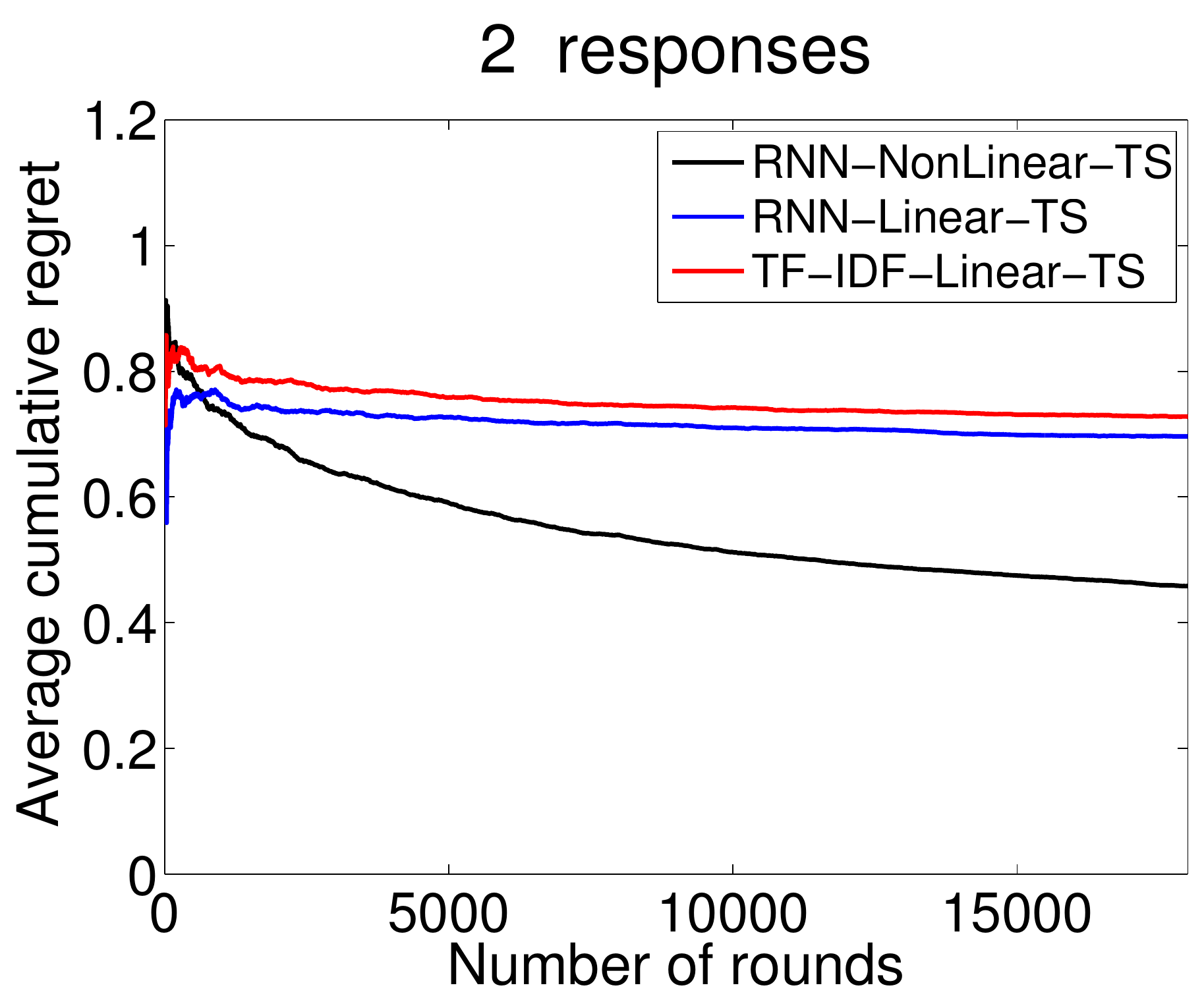}
      \includegraphics[width=0.33\textwidth]{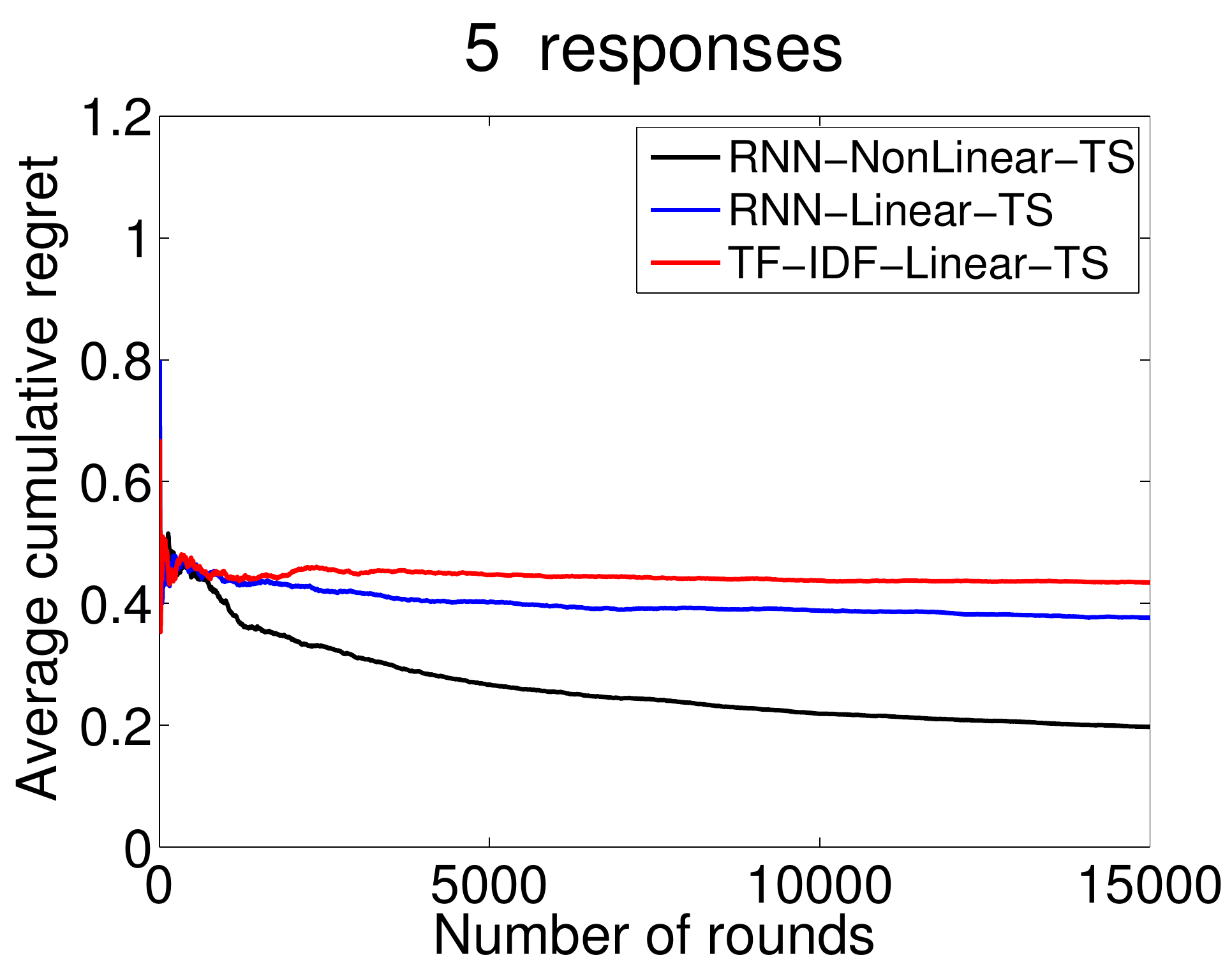}
      \caption{The average cumulative regret along the growing round of interactions in bandits for $k$ ($k$=1, 2, 5) responses returned per interaction. $k$ indicates the number of responses returned by the agent in each interactions. Corresponding rewards for the $k$ responses are provided by the user. Total sample size for online learning is 1000.}
      \label{fig:bandit}
    \end{figure*}

\begin{table*}[ht]
    \centering
    \begin{tabular}{l|c|c|c}
         & {\tt TF-IDF-Linear-TS} & {\tt RNN-Linear-TS} & {\tt RNN-NonLinear-TS}\\\hline
       1 in 10 R@1   & 10.0 \% & 10.5 \%& 24.5 \%\\\hline
       1 in 10 R@2    & 19.5 \% & 21.0 \%& 38.0 \%\\\hline
       1 in 10 R@5     & 60.5 \%& 67.5 \% & 75.0 \%\\
    \end{tabular}
    \caption{Recall@k evaluation results for the proposed bandit models.  {\tt TF-IDF-Linear-TS}, {\tt RNN-Linear-TS} and {\tt RNN-NonLinear-TS} are evaluated when k = $1$, $2$ and $5$.}
    \label{tab:tab}
\end{table*}

\section{Experiments}
    We evaluate the efficiency of the proposed online dialog learning method on the average cumulative regret and response selection Recall@k metrics. We describe the experiment settings and evaluation metrics, and discuss the experimental results.

\subsection{Data Set}
    We use the Ubuntu Dialogue Corpus (UDC)~\cite{lowe2015ubuntu} in our evaluation. UDC is a multi-turn based dialog corpus that is constructed from Ubuntu chat logs, used to receive technical support for Ubuntu-related problems. All named entities in the data set are replaced with corresponding tags (names, locations, organizations, URLs, and paths) in the data pre-processing stage. The data set contains dialog context-response pairs that are extracted and sampled from real chat logs. Each example contains three fields: the dialog context, the candidate response, and a label (match or non-match). 
    
\subsection{Evaluation Metrics}
    We evaluate our model and proposed learning methods on two different metrics, average cumulative regret and Recall@k. The average cumulative regret is defined as: 
    \begin{equation}
    \label{eq:reg}
      R(T) = \frac{\sum_{t = 1}^T r_{t,u_t^{*}} - \sum_{t = 1}^T r_{t,u_t}}{T},
    \end{equation}
    where $r_{t,u_t^{*}}$ is the optimal reward could be achieved at round $t$, and $r_{t,u_t}$ is the reward achieved by bandits at round $t$.

    For the evaluation using Recall@k, the agent is provided with 10 candidate responses. The goal of the agent is to rank the responses by assigning a high score to the true response, and assigning lower scores to the false or distracting responses. Recall@k means that the ranking is considered correct if the true response if among the top $k$ selections make by the agent out of the 10 candidate responses. We report Recall@k with $(n, k) \in \{(10, 1), (10, 2), (10, 5)\}$.

\subsection{Experiment Settings}
    We use a bidirectional LSTM to encode dialog context and response as it has stronger capability in capturing longer term temporal dependencies \cite{hochreiter1997long} comparing to vanilla RNN. LSTM state size and output size are both set as 128. Word embeddings of size 150 are randomly initialized and fine-tuned during mini-batch (size 128) training. We use Adam optimizer \cite{kingma2014adam} in the neural network offline model training with initial learning rate of 1e-3. Dropout \cite{srivastava2014dropout} with keep probability of 0.5 is applied during offline supervised pre-training.

    We compare three different bandit models in the main experimental results: 
    \begin{itemize}
        \item {\tt TF-IDF-Linear-TS}: The context and response features are generated by TF-IDF followed by a dimension reduction operation with PCA. A linear logistic function $\sigma(\bx\bw )$ is used to approximate the reward, where $\bx =[ \bc, \bu]$.
        \item {\tt RNN-Linear-TS}: The dialog context and response are encoded with a bidirectional LSTM. As in {\tt TF-IDF-Linear-TS}, a linear logistic function $\sigma(\bx\bw)$ is used to approximate the reward.
        % $\sigma(u) = \frac{1}{1 + e^{- u}}$
        \item {\tt RNN-NonLinear-TS}: This is our proposed NNBM. The dialog context and response are encoded with a bidirectional LSTM. A nonlinear logistic function $\sigma(\bc\bM\bu^T)$, which is also $\sigma(\phi(\bx)\bw)$, is used to approximate the reward. 
    \end{itemize}

    We further compare our method to Thompson sampling with linear regression proposed in \cite{bouneffouf2014contextual}, as it is the most closely related recent work using bandits with Thompson sampling for dialog learning. Bouneffoutf et al. use a linear bandit with Thompson sampling to handle continuous reward. Our method uses non-linear bandit with Thompson sampling to handle Bernoulli reward. In comparing these two bandit methods, we apply the same text encoding using a bidirectional LSTM. 

    In model training, the bidirectional LSTM encoder is trained using data from the original UDC training set. Online bandit learning and evaluation is performed using data sampled from the UDC test set.

    \begin{figure}[ht]
      \centering
      \includegraphics[width=0.4\textwidth]{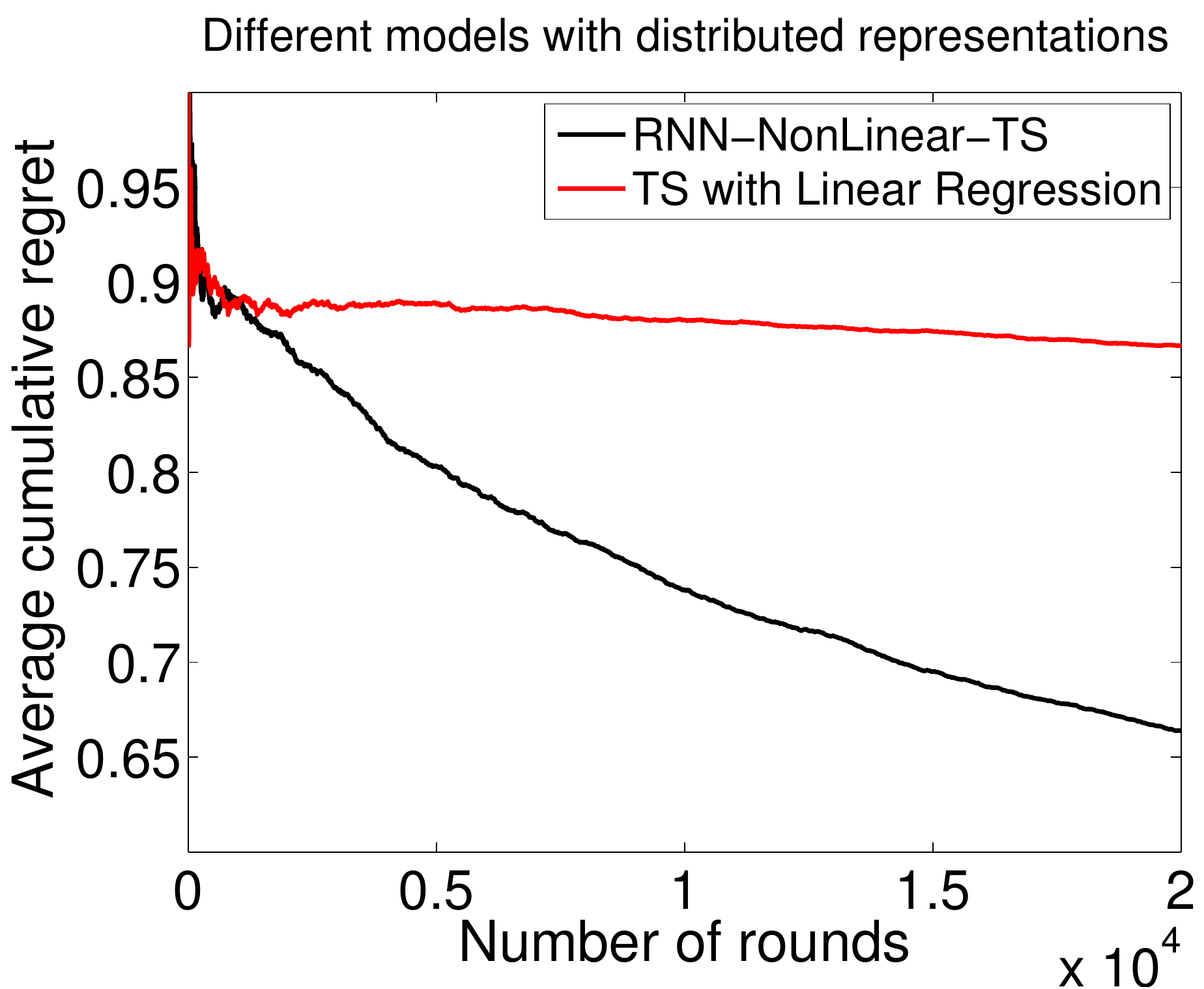}\hspace{-0.05in}
      \caption{Comparison between {\tt RNN-NonLinear-TS} and Thompson sampling with linear regression \cite{bouneffouf2014contextual} in approximating the reward.}
      \label{fig:small}
    \end{figure}

    \begin{table*}[th]
        \centering
        \begin{tabular}{c|c|l}
        \multicolumn{3}{l}{\textbf{Dialog Context:}} \\
         \hline
           User A:   & \multicolumn{2}{l}{what you looking for linuxuz3r? no i mean are you looking for a spefic program?} \\
                & \multicolumn{2}{l}{im not sure if there is anything better then sourceforge} \\
           User B:   & \multicolumn{2}{l}{no particular program, anything that interest me then contribute to the source. }\\
                & \multicolumn{2}{l}{i wanna learn how to read code} \\\hline
         %\multicolumn{3}{c}{} \\
          \multicolumn{3}{c}{} \\
        \multicolumn{3}{l}{\textbf{Interaction 1:}} \\
           \textbf{Rank} & \textbf{Confidence} & \textbf{Response} \\\hline
           1 & 0.58374  & other then checking the addational-drivers tool (jockey-gtk) and see if you \\
             &     & got the drivers installed..  thats all i know about ati cards. \\
           2 & 0.46708  & we needed some custom features not all the features of ffmpeg so i needed  \\
             &     & those features only. \\
           3 & 0.37162  & good lad \\\hline

        \multicolumn{3}{c}{} \\
        \multicolumn{3}{l}{\textbf{Interaction 4:}} \\
    
           \textbf{Rank} & \textbf{Confidence} & \textbf{Response} \\\hline
           1 & \textbf{0.62103}  & \textbf{there is one that escapes me at the moment. most people use sourceforge} \\
           2 & 0.46708  & possible \\
           3 & 0.46526  & we needed some custom features not all the features of ffmpeg so i needed  \\
             &     & those features only. \\\hline
    
        \multicolumn{3}{c}{} \\
        \multicolumn{3}{l}{\textbf{Interaction 7:}} \\
    
           \textbf{Rank} & \textbf{Confidence} & \textbf{Response} \\\hline
           1 & \textbf{0.87181}  & \textbf{there is one that escapes me at the moment. most people use sourceforge} \\
           2 & 0.17525  & giles that is the exact command, replace username with the actual name of the \\
             &     & account you are trying to change. ex. 'sudo passwd kriskropd' \\
           3 & 0.10332  & we needed some custom features not all the features of ffmpeg so i needed  \\
           &     & those features only. \\\hline
        \end{tabular}
        \caption{An example showing how bandit model learns with increasing number of interactions. In each round of interactive learning, a context is randomly selected as input. Here we show the response ranking to one particular context sample. Interaction 1, 4, 7 indicates the round of interactions when this particular context appears for the $1$st, $4$th and $7$th time respectively. For each response, logistic regression model in bandits provides a confidence score, based on which we sort all the response and show the top three. The response in bold represents the true response. }
        \label{tab:example}
    \end{table*}

\subsection{Online Learning Evaluations}
    We evaluate the performance of different bandit models in the online setting. We select $1000$ samples from the UDC test set for the online model evaluation. In each round of interaction,  a context is randomly selected from the $1000$ samples and provided to the agent. Given the context, the agent returns $k$ responses to the user. The user then provides a binary (1 or 0) feedback on each of the returned response, where $1$ indicates a good response, and $0$ indicates a bad response. If a bad response is selected, the regret value, which is initialized to 0, increments by $1$. We report the average cumulative regret in Equation \ref{eq:reg} over $50000$ rounds of interactions.

    \textbf{RNN-Linear-TS versus TF-IDF-Linear-TS.} The results are reported in Figure \ref{fig:bandit}. As can be observed from these figures, with increasing number of rounds of interactions in bandit model learning, the average cumulative regret gradually decreases and finally converges. Comparing to RNN based text encoding method, TF-IDF as a count-based method that runs faster and requires much less computation. However, it ignores the word ordering and does not learn word relations in the sentence. For bandits using linear logistic function, {\tt RNN-Linear-TS} outperforms {\tt TF-IDF-Linear-TS} in all experiments. This shows the advantages of using RNN methods for test feature representation that captures long term temporal dependencies in text comparing to conventional methods using TF-IDF. 

    \textbf{RNN-NonLinear-TS versus RNN-Linear-TS.}  As illustrated in Figure \ref{fig:bandit}, {\tt RNN-NonLinear-TS} beats {\tt RNN-Linear-TS} by large margin at convergence in all experiment settings, at a cost of higher computational complexity. The proposed nonlinear logistic regression Thompson sampling method produces much better results in this online dialog response selection task, compared to the linear models that are widely applied in online advertisements. With total sample size of 1000 (Figure \ref{fig:bandit}) and a single returned response (i.e. $k=1$), the average cumulative regret converges at $0.61$, indicating the probability of bandits selecting correct response is $0.39 (=1-0.61)$ in each round.

    We further compare our {\tt RNN-NonLinear-TS} methods to previous work on using linear regression \cite{bouneffouf2014contextual} for reward approximation in dialog systems. In comparing these two bandit methods, we used the same text encoding by LSTM. As illustrated in Figure \ref{fig:small}, our proposed NNBM method demonstrates strong advantage in online learning efficiency comparing to previously proposed method.

    Table \ref{tab:example} shows an example showing how bandit model learns with increasing number of online learning interactions. It can be seen that with more rounds of interactions, the model confidence on the true response increases and the gap between the top two confidence scores expands.

    \textbf{On number of responses.} We experiment with letting the agent to select different number of responses for each given dialog context, with $k$ value being 1, 2 and 5. The results are shown in Figure \ref{fig:bandit}. As expected, the true response is more likely to appear in the returned responses when the number of candidate responses $k$ increases, and thus resulting in a lower average cumulative regret. The gap between {\tt RNN-NonLinear-TS} and {\tt RNN-Linear-TS} expands when $k$ grows from $1$ to $2$, and shrinks when $k$ grows further to 5. This might be explained by the power of {\tt RNN-NonLinear-TS} model in ranking the true response higher among the candidate responses. 

\subsection{Recall@k Results}
    We further evaluate the performance of our dialog response selection model using Recall@k. We split our $1000$ learning samples into two parts: $800$ samples are used in the online learning by bandits, and the rest $200$ samples are set aside for evaluation. The results are shown in Table \ref{tab:tab}. Note that these numbers should not be directly compared to the ones in \cite{lowe2015ubuntu} and \cite{kadlec2015improved}, which are reported under offline model training settings using the entire training set. These Recall@k evaluation results are promising given the very small size ($800$) of samples involved in the online model learning. As in the above online evaluation results, {\tt RNN-Linear-TS} outperforms {\tt TF-IDF-Linear-TS} for all $k$ values under the Recall@k metric. {\tt RNN-NonLinear-TS} consistently outperforms {\tt RNN-Linear-TS} by a large margin. 
    
\section{Conclusions}
    In this paper, we proposed a neural nonlinear bandit model (NNBM) using distributed representations of text in contextual multi-armed bandits for online response selection in dialog modeling. We designed a customized Thompson sampling method that is applied to a polynomial feature space in approximating the reward for bandits. Our experiments results on Ubuntu Dialogue Corpus showed clear advantages of using distributed representation of text produced by neural network methods in learning text representations comparing to using TF-IDF method for online dialog model learning. The experimental results demonstrated significant performance gain of the proposed methods over conventional linear contextual bandits, which are widely used in online advertisement and recommendations. We also reported encouraging Recall@k evaluation results of NNBM in best response selection. 

    To our best knowledge, this is the first attempt at combining neural network methods and contextual multi-armed bandits in the dialog response selection online learning setting. We believe the proposed online learning method can be very useful in situations where the size of labeled data set is limited.
    
\section{Acknowledgments}
This work is partially supported by Adobe donation. We thank Branislav Kveton for the helpful discussions.

\bibliographystyle{aaai}
\bibliography{reference}

\end{document}